\documentclass[conference]{IEEEtran}
\IEEEoverridecommandlockouts

\usepackage{cite}
\usepackage{amsmath,amssymb,amsfonts}
\usepackage{algorithmic}
\usepackage{graphicx}
\usepackage{textcomp}
\usepackage{subcaption}  
\usepackage{xcolor}
\usepackage{orcidlink}
\def\BibTeX{{\rm B\kern-.05em{\sc i\kern-.025em b}\kern-.08em
    T\kern-.1667em\lower.7ex\hbox{E}\kern-.125emX}}
\usepackage{tikz}
\usetikzlibrary{shapes, arrows, positioning}

\begin{document}

\title{Diffusion-Driven Inertial Generated Data for Smartphone Location Classification}

\author{
\IEEEauthorblockN{
Noa Cohen\IEEEauthorrefmark{1}\thanks{*Corresponding author: ncohe158@campus.haifa.ac.il}, 
Rotem Dror\IEEEauthorrefmark{2}, 
Itzik Klein\IEEEauthorrefmark{1}~\orcidlink{0000-0001-7846-0654}
}
\IEEEauthorblockA{\IEEEauthorrefmark{1}The Hatter Department of Marine Technologies, University of Haifa, Israel}
\IEEEauthorblockA{\IEEEauthorrefmark{2}The Department of Information Systems, University of Haifa, Israel}
}

\maketitle

\begin{abstract}
Despite the crucial role of inertial measurements in motion tracking and navigation systems, the time-consuming and resource-intensive nature of collecting extensive inertial data has hindered the development of robust machine learning models in this field. In recent years, diffusion models have emerged as a revolutionary class of generative models, reshaping the landscape of artificial data generation. These models surpass generative adversarial networks and other state-of-the-art approaches to complex tasks. In this work, we propose diffusion-driven specific force-generated data for smartphone location recognition. We provide a comprehensive evaluation methodology by comparing synthetic and real recorded specific force data across multiple metrics. Our results demonstrate that our diffusion-based generative model successfully captures the distinctive characteristics of specific force signals across different smartphone placement conditions. Thus, by creating diverse, realistic synthetic data, we can reduce the burden of extensive data collection while providing high-quality training data for machine learning models.
\end{abstract}

\section{Introduction}
\noindent Indoor navigation systems have become increasingly essential in modern society, addressing critical needs in environments where global navigation satellite systems (GNSS) signals are unavailable\cite{ zafari2019survey, yassin2016recent}. These systems enable efficient wayfinding in complex structures such as shopping malls, airports, and hospitals, while supporting emergency response operations, asset tracking in warehouses, and accessibility assistance for individuals\cite{davidson2016survey}. With the rapid expansion of smart infrastructure and mobile computing capabilities, the demand for accurate and reliable indoor positioning continues to grow in numerous applications, including location-based services, personalized marketing, facility management, and smart building operations.
\noindent
Pedestrian dead reckoning (PDR) has emerged as a promising approach for indoor positioning that utilizes inertial sensor measurements to track a user's movements \cite{klein2025pedestrian, hou2020pedestrian}. By continuously measuring specific force and angular velocity through accelerometers and gyroscopes, PDR systems estimate displacement and direction changes to determine a user's position relative to a known starting point. This approach is particularly valuable as it operates independently of external infrastructure, offering a self-contained solution that functions across diverse indoor environments without requiring additional hardware installation or maintenance.
\noindent
Human activity recognition (HAR) and smartphone location recognition (SLR) play a crucial role in enhancing PDR systems and enabling numerous other applications \cite{klein2019smartphone, chen2021deep, wu2021adaptive ,wang2021three}. HAR identifies specific user activities (walking, running, climbing stairs), allowing PDR algorithms to apply appropriate motion models and reduce positioning errors \cite{elhoushi2015motion}. SLR determines where a device is carried (hand, pocket, bag), which significantly impacts the characteristics of sensor readings and subsequent motion estimation accuracy. Beyond PDR, these technologies support healthcare monitoring \cite{nweke2018deep}, fitness tracking \cite{ravi2016deep}, and context-sensitive computing \cite{yao2017deepsense}.
\noindent
HAR and SLR algorithms employ machine learning (ML) and deep learning (DL) techniques. ML methods employ feature engineering with classifiers like support vector machines and decision trees \cite{alam2024neurohar}. Recently, most works employ DL approaches such as convolutional neural networks (CNNs) \cite{ronao2016human}, recurrent neural networks (RNNs) \cite{hammerla2016deep}, hybrid architectures \cite{ordonez2016deep}, and transformers\cite{shavit2021boosting} to automatically extract temporal and spatial features from raw inertial data. Despite the advancement of DL architectures for high-performance HAR and SLR, most works were evaluated on a limited amount of participants, as lack of high-quality training data remains a significant barrier. Deep networks, with their numerous parameters, require vast and diverse datasets to achieve optimal results and generalizability. However, collecting extensive inertial data is a time-consuming, resource intensive process, and requires diverse participant recruitment to ensure representative data across different humans and movement patterns.
\noindent
In recent years, diffusion models \cite{ho2020denoising, song2019generative} have emerged as a revolutionary class of generative models, reshaping the landscape of artificial data generation. These models have surpassed GANs in complex tasks such as image synthesis \cite{ croitoru2023diffusion} and are expanding across various domains \cite{yang2023diffusion, kong2020diffwave}. For time series generation, approaches like CSDI \cite{tashiro2021csdi} and autoregressive diffusion denoising models \cite{rasul2021autoregressive} have shown promising results by operating directly in the time domain. Recent work by Naiman et al. \cite{naiman2024utilizing} demonstrated the transformation of time series into images using delay embedding and a short-time Fourier transform before applying diffusion models. In inertial sensing, recent diffusion-based approaches \cite{shao2023study, oppel2024imudiffusion} typically operate directly on time-domain signals rather than leveraging the transformation to image representations.
\noindent
In this work, we propose diffusion-driven specific force generated data for smartphone location recognition.  We explore the transformation of accelerometer time series measurements into image representations by delay embedding before applying diffusion processes, as suggested in \cite{naiman2024utilizing}. This transformation allows us to leverage the power of well-established vision-based diffusion models \cite{rombach2022high} while preserving the essential temporal characteristics of the original signals. By conditioning the generation process on device placement labels (hand, bag, body, leg), our method produces diverse, realistic, and class-specific inertial datasets that capture the distinct motion patterns associated with different smartphone carrying positions. The key contributions of our work include:
 \begin{enumerate}
    \item A novel framework for specific force measurement data generation using delay embedding and conditional diffusion models
    \item  A comprehensive evaluation methodology comparing synthetic and real specific force data across multiple metrics.
\end{enumerate}
Our approach has the potential to significantly reduce the burden of inertial data collection while maintaining or even improving data quality and relevance for SLR applications. The successful implementation of our image-based diffusion model for specific force data generation could have far-reaching implications. It could accelerate the development of more robust and accurate inertial -based systems, enable more efficient and effective training of ML/DL models in this domain, and potentially extend to other sensor-based applications in fields such as robotics, healthcare, and human-computer interaction.
\noindent
The remainder of this paper is organized as follows: Section \uppercase\expandafter{\romannumeral 2} presents our proposed approach, detailing the signal-to-image transformation using delay embedding, the diffusion model architecture, and our evaluation methodology. Section \uppercase\expandafter{\romannumeral 3} describes our experimental setup and presents comprehensive results, with analyses of the synthetic data quality through visual, statistical, and classifier-based evaluations. Finally, Section \uppercase\expandafter{\romannumeral 4} concludes this paper.
\section{Proposed Approach}
\label{sec:Proposed Approach}
\noindent Our approach to generating synthetic specific force data is inspired by the work of Naiman et al. \cite{naiman2024utilizing}, who demonstrated the effectiveness of transforming time series into image representations before applying diffusion models. We adapt this methodology specifically for specific force signals, leveraging the strengths of vision-based diffusion architectures while preserving the temporal characteristics essential for realistic motion patterns.

\begin{figure*}[t]
  \centering
  \includegraphics[width=\textwidth]{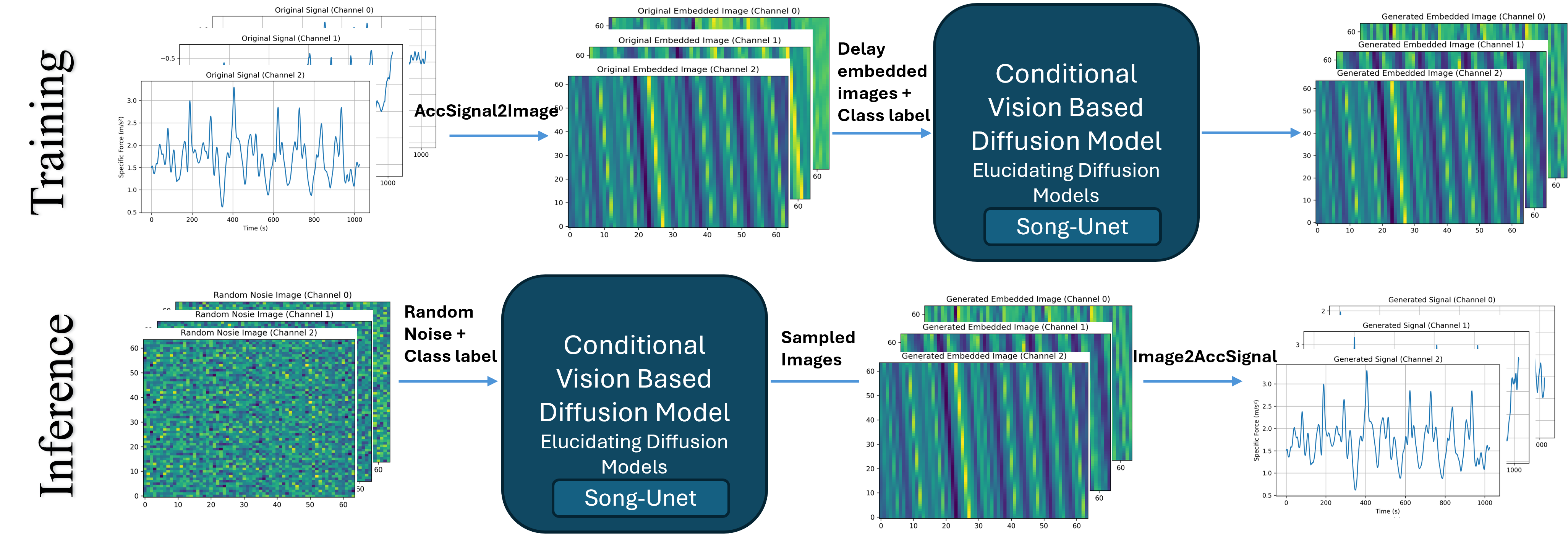}
  \caption{Block diagram of our proposed approach for synthetic specific force signal generation, showing the three main stages: signal-to-image transformation, using the AccSignal2Image transformation, a vision-based diffusion model implementing elucidating diffusion models with Song-Unet as neural backbone, and image-to-signal reconstruction, using the Image2AccSignal transformation.}
  \label{fig:pipeline}
\end{figure*}

 We implement a three-stage pipeline (Figure~\ref{fig:pipeline}) consisting of: 1) a signal-to-image transformation module using delay embedding, 2) a vision-based diffusion model for generating high-quality embedded representations, and 3) an image-to-signal transformation module to reconstruct synthetic time series. In the following section, we elaborate on each part of our propose approach.

\subsection{Signal-to-Image Transformation}
The first stage of our pipeline transforms raw specific force signals into image representations using the delay embedding technique \cite{naiman2024utilizing}. Given an input time series $\mathbf{x} \in \mathbb{R}^{L \times 3}$ with $L$ timesteps and 3 features (specific force channels), we transform it to an image representation $\mathbf{x}_{\text{img}} \in \mathbb{R}^{C \times H \times W}$ using delay embedding with parameters $\tau = 15$ (time delay) and $n = 64$ (embedding dimension).
Delay embedding \cite{naiman2024utilizing} is an invertible transformation that arranges segments of the signal into columns of a matrix. For a univariate time series $\mathbf{x}_{1} \in \mathbb{R}^L$, with user parameters $m$ (skip value) and $n$ (column dimension), we construct the matrix:
\begin{equation}
\mathbf{X} = 
\begin{bmatrix}
x_1 & x_{m+1} & \cdots & x_{L-n} \\
\vdots & \vdots & \ddots & \vdots \\
x_n & x_{n+m+1} & \cdots & x_L
\end{bmatrix} \in \mathbb{R}^{n \times q}
\end{equation}

\begin{figure}[t]
\centering
\includegraphics[width=\linewidth]{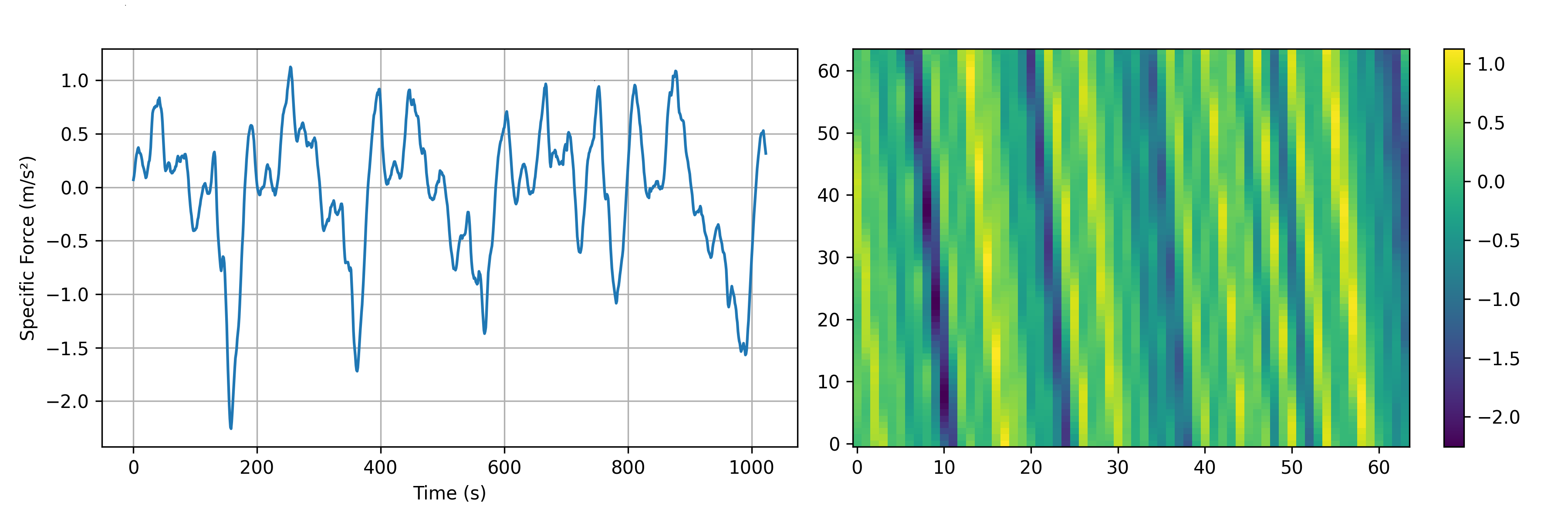}
\caption{(left) Original recorded accelerometer x-axis readings. (right) The resulting delay embedded image using AccSignal2Image transformation.}
\label{fig}
\end{figure}

$q = \lceil(L - n)/m\rceil$. The image $\mathbf{x}_{\text{img}}$ is created by padding with zeros to fit the neural network input constraints. This transformation is invertible, allowing reconstruction of the original signal from the generated image. For instance, when $m = 1$, the original time series can be recovered by concatenating the first row and last column of the image. We refer to this specific force signal-to-image transformation process as AccSignal2Image. The AccSignal2Image transformation effectively captures the temporal dynamics of accelerometer data by organizing sequential relationships into a spatial format that convolutional neural networks can efficiently process. This transformation preserves the rich information contained in the raw specific force signals while presenting it in a format optimized for deep learning architectures.\\
\noindent
A key advantage of the delay embedding transformation is its reconstructive invertibility, meaning that the original specific force signals can be precisely recovered from their embedded image representations. We denote this reverse operation as Image2AccSignal, which systematically reverses the embedding process, reconstructing the specific force signal by rearranging image segments according to the known embedding parameters ($\tau$ and $n$). \\
\noindent
For our specific force signals with multiple channels, we apply this transformation to each channel independently and combine them into a multi-channel image representation. This creates a 2D representation where temporal patterns are encoded as spatial relationships (see Figure~\ref{fig}).
\subsection{Diffusion Model Architecture}
Our generative framework adopts the Elucidated Diffusion Models (EDM) architecture proposed by Karras et al.~\cite{karras2022elucidating}, which systematically reorganizes the design space of diffusion-based generative models. Unlike earlier models, EDM provides a modular and empirically optimized setup that decouples training and sampling procedures, enabling significant gains in quality and efficiency. The formulation of the generative process is based on the probability flow ordinary differential equation (ODE), where the noise schedule $\sigma(t)$ governs the desired noise level at any given time $t$. This noise schedule controls the rate at which noise is added or removed from the sample during the forward or backward evolution, respectively. The time derivative of the noise schedule, $\dot{\sigma}(t)$, captures the rate of change of the noise level with respect to time. The ODE governing this evolution is given by \cite{karras2022elucidating}:
\begin{equation}
d\mathbf{x} = -\dot{\sigma}(t){\sigma(t)} \nabla_\mathbf{x} \log p(\mathbf{x}; \sigma(t)) \, dt
\end{equation}
where a sample $\mathbf{x}_a \sim p(\mathbf{x}; \sigma(t_a))$ evolves from time $t_a$ to time $t_b$, resulting in $\mathbf{x}_b \sim p(\mathbf{x}; \sigma(t_b))$.
\noindent
The score function $\nabla_\mathbf{x}\log p(\mathbf{x}; \sigma)$ is approximated by a neural denoiser $D_\theta(\mathbf{x}, \sigma)$ preconditioned according to EDM's parameterization: 
\begin{equation} 
D_\theta(\mathbf{x}, \sigma, \mathbf{l}) = c_{skip}(\sigma)\mathbf{x} + c_{out}(\sigma)\mathbf{F}_\theta(c_{in}(\sigma)\mathbf{x}, c_{noise}(\sigma), \mathbf{l}) 
\end{equation} 
\noindent
Here, $\mathbf{F}_\theta$ represents the core neural network architecture (typically a U-Net) that performs the denoising operation, while $\mathbf{l}$ denotes the class-conditioning vector that guides the model toward generating samples from a specific class. The preconditioning coefficients are carefully defined to optimize the training dynamics across different noise levels:
\begin{IEEEeqnarray}{rCl}
c_{skip}(\sigma) &=& \frac{\sigma_{data}^2}{\sigma^2+\sigma_{data}^2}, \quad
c_{out}(\sigma)=\frac{\sigma\,\sigma_{data}}{\sqrt{\sigma^2+\sigma_{data}^2}}, \nonumber\\[1mm]
c_{in}(\sigma)   &=& \frac{1}{\sqrt{\sigma^2+\sigma_{data}^2}}, \quad
c_{noise}(\sigma)=\frac{\ln(\sigma)}{4}
\end{IEEEeqnarray}
where $c_{skip}(\sigma)$ modulates the skip connection, $c_{in}(\sigma)$ and $c_{out}(\sigma)$ scale the input and output magnitudes respectively, and $c_{noise}(\sigma)$ maps noise level $\sigma$ into a conditioning input for $\mathbf{F}_\theta$. \\
\noindent
The backbone network $\mathbf{F}_\theta$ is implemented as SongUNet~\cite{song2020score}, which integrates residual blocks and multi-resolution attention, making it particularly effective for structured signal representations. The conditioning variable $c$ reflects class labels, such as sensor placement, enabling conditional generation. \\
\noindent
In line with the EDM setup, we use a log-normal noise distribution for training, with $\sigma_{min} = 0.002$, $\sigma_{max} = 80$, and $\sigma_{data} = 0.5$. Sampling is performed over $T = 18$ steps using Heun's second-order method, which balances accuracy and speed. This setup enables us to achieve high sample quality with significantly fewer function evaluations compared to previous diffusion models.
\noindent
For the network architecture, we configured the UNet with channel multipliers [1, 2, 2, 2], model channels of 64, and attention mechanisms at $16 \times 16$ resolution, striking a balance between computational efficiency and model expressiveness.
\subsection{Diffusion Model Training Process}
The training process for our diffusion model begins with the already transformed image representations of specific force signals from the delay embedding step. Given delay-embedded image representations of specific force signals, we train the model to predict a transformed target using a parameterization dependent on $\sigma$ that balances the magnitudes of the gradients at different noise levels.
The loss function is defined as \cite{karras2022elucidating}:
\begin{align}
L(\theta) &= \mathbb{E}_{\sigma, y, n} \Big[\lambda(\sigma) 
    \big\|F_{\theta}\big(c_{\text{in}}(\sigma)(y + n), c_{\text{noise}}(\sigma)\big) \nonumber\\
    &\qquad\qquad - \text{target}_{\sigma}(y, n)\big\|^2 \Big]
\end{align}
where $F_\theta$ is the trainable network, $n \sim \mathcal{N}(0, \sigma^2 I)$, and the effective target is derived from the underlying reconstruction formula:
\begin{equation}
\text{target}_{\sigma}(y, n) = \frac{1}{c_{\text{out}}(\sigma)}(y - c_{\text{skip}}(\sigma))(y + n)
\end{equation}
To balance the effective per-sample contribution to the loss, we set the weight $\lambda(\sigma) = 1 / c_{\text{out}}(\sigma)^2$.The noise level $\sigma$ for each training sample is drawn from a log-normal distribution $p_{\text{train}}(\sigma)$. The result of this training process is a neural network capable of accurately modeling the conditional distributions of the embedded images at varying noise levels. Consequently, during inference, we can effectively sample and reconstruct realistic synthetic specific force signals corresponding to distinct placement classes. Specifically, we sample noise from a standard normal distribution, specify the desired placement class through conditioning, and apply the reverse diffusion process to obtain a clean image. We then transform this image back to the time domain using the inverse delay embedding transformation to produce synthetic specific force signals with characteristics specific to the chosen placement class.

\subsection{Smartphone location recognition}
\label{subsec:Smartphone location recognition}
To systematically evaluate the quality of our synthetic specific force data, we implemented a cross-evaluation framework using CNNs. We designed two distinct classifiers with the goal of assessing how well our synthetic data matches the statistical distribution of the real data.

\paragraph{Image-based CNN Classifier} Our first classifier operates directly on the delay-embedded images from specific force signals. The network architecture consists of four convolutional layers with 16, 32, 64, and 128 filters, respectively, each with a kernel size of $3 \times 3$ and stride of 1. Each convolutional layer is followed by batch normalization, ReLU activation, and $2 \times 2$ max-pooling. These layers are followed by an adaptive average pooling layer that produces a fixed-size feature map regardless of input dimensions. The features are then flattened and passed through a fully-connected layer with 256 neurons and ReLU activation, followed by a dropout layer (rate = 0.5) for regularization. The final layer consists of N output neurons with softmax activation for classifying each of the N placement categories.

\paragraph{Signal-based CNN Classifier} Our second classifier works with reconstructed specific force signals in the time domain rather than their image representations. The architecture mirrors the image-based classifier but replaces 2D operations with their 1D counterparts. Specifically, it uses four 1D convolutional layers with the same filter counts (16, 32, 64, 128), kernel size of 5, and stride of 1. Each convolutional layer is followed by batch normalization, ReLU activation, and 1D max-pooling with a pool size of 2. The subsequent layers maintain the same structure as the image-based classifier, with adaptive pooling, flattening, a fully-connected hidden layer, dropout, and a N-neuron output layer with softmax activation.

\section{Experiments and Results}
\label{sec:experiments}

\noindent In this section we present our extensive experiments made to evaluate our approach on the robust IMU double integration (RIDI) dataset \cite{yan2018ridi}, which contains specific force readings from smartphones in various placements.

\subsection{Dataset and Preprocessing}
The RIDI dataset encompasses over 100 minutes of motion data at 200Hz from six human subjects, featuring four distinct smartphone placements: in a leg pocket, in a bag, hand-held, and body-mounted. It captures various motion patterns including forward/backward walking, side movements, and acceleration/deceleration sequences. The dataset also includes a preprocessing step where asynchronous signals from various sources were synchronized to the Tango pose timestamps using linear interpolation.
\noindent Our preprocessing pipeline includes:
\begin{enumerate}
    \item Extracting triaxial specific force data (x, y, z channels).
    \item Removing initial stabilization artifacts by discarding the first 1500 samples.
    \item Segmenting signals into fixed-length windows of 1024 samples with 50\% overlap.
    \item Channel-wise normalization was applied using the mean and standard deviation computed independently for each channel of the dataset.
\end{enumerate}

\noindent After the preprocessing phase, the EDM model was trained using the AdamW optimizer \cite{loshchilov2017decoupled} with a learning rate of $10^{-4}$ and batch size of 128 for 1000 epochs.

The two classifiers, defined in Section ~\ref{subsec:Smartphone location recognition}, were implemented in PyTorch and trained exclusively on real data using an Adam optimizer with a learning rate of 0.001, weight decay of $10^{-5}$, and cross-entropy loss. We employed early stopping with a patience of 10 epochs to prevent overfitting, and used a training/validation/test split of 70\%/15\%/15\%.
\subsection{Visual Distribution Analysis}
\noindent
Once the training was completed, we randomly generated class labels and sampled from the trained diffusion model to produce synthetic data. Below, we present a through analysis and evaluation of our generated data using different metrics and approaches
\subsubsection{ t-distributed stochastic neighbor embedding } We employed t-distributed stochastic neighbor embedding (t-SNE) \cite{der2008visualizing} to visualize the high-dimensional distribution of both real and synthetic specific force data. Figure \ref{fig:three_channels} presents the visualization across all three specific force data channels (x, y, and z). 
For each channel, we observed substantial overlap between the real and synthetic distributions, with similar cluster formations across all four placement classes. The embeddings clearly demonstrate that our synthetic data accurately captures the class-specific characteristics, with well-defined boundaries between different placement types (bag, body, handheld, and leg). This visual evidence suggests that our diffusion model successfully learned the distinctive motion patterns associated with each smartphone placement.
\subsubsection{Probability density functions} 
\begin{figure*}[t]
\centering
\includegraphics[width=\linewidth]{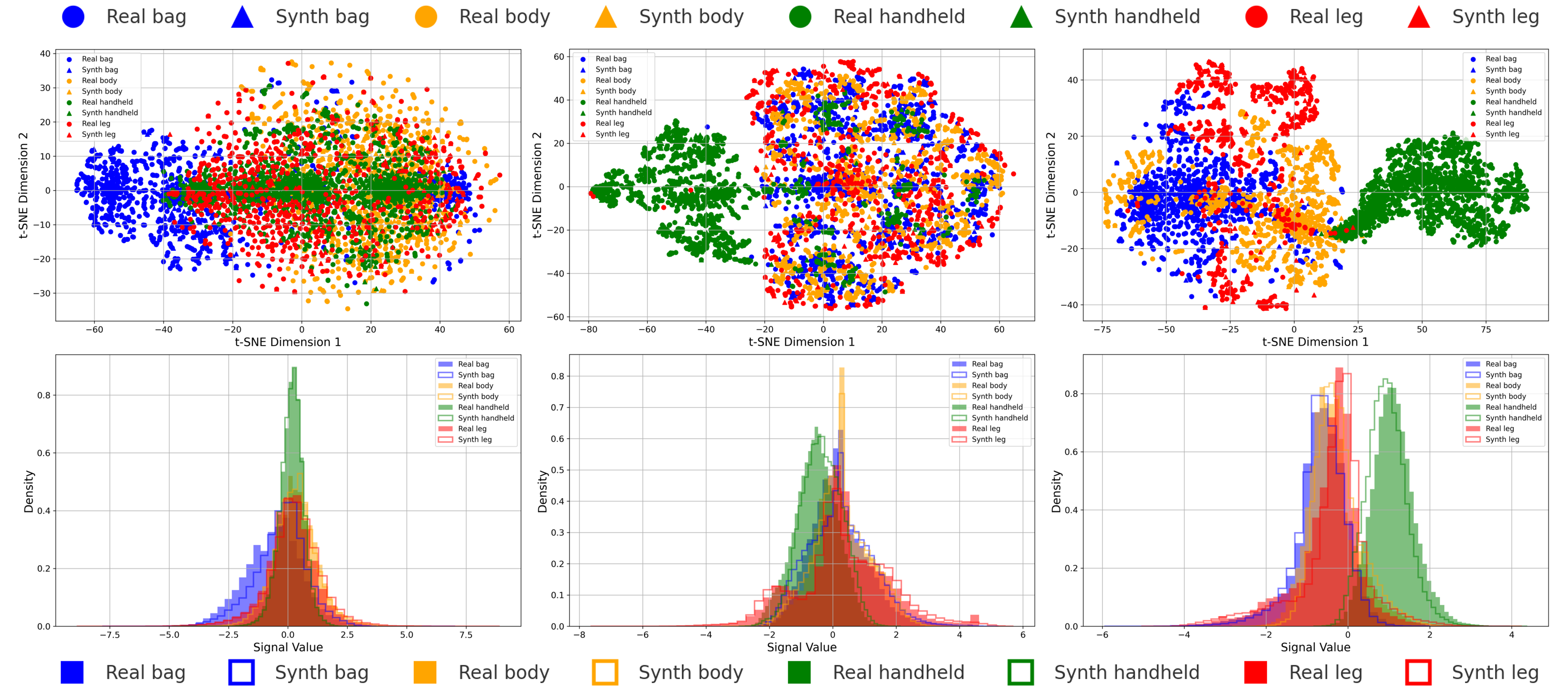}
\caption{Comparison between real and synthetic data. Top row: t-SNE embeddings of the real and generated signals for each specific-force channel (x, y and z). Bottom row: Corresponding probability density functions illustrating the distributional similarity between real and synthetic data. The legends at the top and bottom differentiate between real and synthetic signals across various placements.}
\label{fig:three_channels}
\end{figure*}

The bottom panels of Figure \ref{fig:three_channels} display the probability density function (PDF) comparisons between real and synthetic data for each specific force data channel. The PDF analysis reveals that our synthetic data not only matches the central tendencies of the real distributions but also accurately reproduces the characteristic peaks, spreads, and tails specific to each channel. 
\subsection{Fréchet inception distance}
\noindent
To quantitatively assess the quality of our generated image samples, we calculated the Fréchet Inception Distance (FID) \cite{ho2022cascaded} between real and synthetic distributions.  The FID computation involved extracting features from both real and synthetic samples using an InceptionV3 network adapted for our delay-embedded images. We then calculated the Fréchet distance between the multivariate Gaussian distributions fitted to these feature spaces. Our approach achieved an FID score of 1.22, which is remarkably low for complex time-series data. For comparison, recent work on RF-based signal generation for pose estimation reported an FID of 1.42 for synthetic RF signals, with a baseline of 0.73 for real data~\cite{zhang2023rf}. Similarly, EEG-to-image generation using diffusion models yielded FID scores of 243.88 in early configurations, later improved to 41.62 using a convolutional conditional model~\cite{eegdiff2024}. These comparisons emphasize that our FID of 1.22 reflects excellent alignment between the generated and real image distributions, confirming both the visual fidelity and deep structural accuracy of the synthetic IMU samples.
\subsection{Classifiers elevation}
\noindent
Using the dual-classifier framework described in Section~\ref{subsec:Smartphone location recognition}, we evaluated how well our synthetic data preserves the distinctive characteristics of different smartphone placement patterns. Both classifiers were trained exclusively on real data and then evaluated on both real test data and synthetic data generated by our diffusion model. Table~\ref{tab:classification_results} summarizes the classification performance across both models and data types.
\begin{table}[h]
\centering
\caption{Classification accuracy (\%) of CNN models trained on real data and evaluated on both real and synthetic test sets.}
\label{tab:classification_results}
\begin{tabular}{lcc}
\hline
\textbf{Model} & \textbf{Real Test Data} & \textbf{Synthetic Data} \\
\hline
Image-based CNN & 97.90\% & 97.40\% \\
Signal-based CNN & 98.13\% & 97.20\% \\
\hline
\end{tabular}
\end{table}
Our image-based CNN classifier achieved 97.9\% accuracy on the real test set and maintained 97.4\% accuracy on synthetic samples, showing a minimal performance gap of only 0.6\%. The signal-based CNN classifier performed similarly well, achieving 98.13\% accuracy on real test data and 97.2\% on synthetic signals, with a small gap of 0.93\%.
These results provide strong evidence that our diffusion-based generative model successfully captures the distinctive characteristics of specific force signals across different smartphone placement conditions. The high accuracy on synthetic data across both classification domains (image and signal) indicates that our model preserves both the spatial features in delay-embedded representations and the temporal dynamics in the reconstructed signals. 
\section{Conclusion}
\label{Conclusion}
\noindent In this study, we have demonstrated the effectiveness of diffusion models for generating synthetic specific force data. By transforming time-series signals into image representations through delay embedding, we were able to leverage the power of vision-based diffusion models to generate high-quality synthetic samples that closely match the statistical properties and functional utility of real specific force data.
Our approach achieved impressive results, with CNN classifiers showing comparable performance on synthetic data compared to real data. The high-quality t-SNE visualizations and low FID scores further confirm the fidelity of our synthetic samples. This high-fidelity generation is particularly valuable for applications requiring realistic sensor
data, such as activity recognition, pedestrian dead reckoning,and device position detection.
The success of our method has significant implications for addressing data scarcity in inertial-based applications. By generating diverse, realistic synthetic data, we can reduce the burden of extensive data collection while still providing high-quality training data for machine learning models.

\bibliographystyle{IEEEtran}
\bibliography{references}

\end{document}